\documentclass{article}

\PassOptionsToPackage{numbers,sort}{natbib}
\usepackage{wrapfig}
 \usepackage[preprint]{neurips_2026}

\usepackage[utf8]{inputenc} 
\usepackage[T1]{fontenc}    
\usepackage{hyperref}       
\usepackage{url}            
\usepackage{booktabs}       
\usepackage{amsfonts}       
\usepackage{nicefrac}       
\usepackage{microtype}      
\usepackage{xcolor}         
\usepackage{graphicx}
\usepackage{tikz}
\usepackage{bbm}
\usepackage{amsmath}
\usepackage{caption}
\definecolor{hotpink}{rgb}{1, 0.41, 0.71}

\usepackage{listings}
\usepackage{xcolor}
\lstdefinestyle{prompt}{
basicstyle=\ttfamily\footnotesize,
breaklines=true,
breakatwhitespace=true,
columns=fullflexible,
keepspaces=true,
showstringspaces=false,
frame=single,
framesep=6pt,
rulecolor=\color{black!30},
backgroundcolor=\color{black!3},
xleftmargin=4pt,
xrightmargin=4pt,
aboveskip=8pt,
belowskip=8pt,
upquote=true,
literate={--}{{-{}-}}1
}

\usetikzlibrary{shadows} 

\title{Masked Visual Actions for Unified World Modeling}

\author{%
  Hadi Alzayer$^{1,2}$ \\
  \And
  Wenlong Huang$^1$ \\
  \And
  Haonan Chen$^{1,3}$ \\
  \And
  Christopher Luey$^1$ \\
  \AND
  Lvmin Zhang$^1$ \\
  \And
  Maneesh Agrawala$^1$ \\
  \And
  Gordon Wetzstein$^1$ \\
  \And
  Li Fei-Fei$^1$ \\
  \AND
  Yilun Du$^3$ \\
  \And
  Jiajun Wu$^1$ \\
  \And
  Jia-Bin Huang$^2$ \\
  \AND
  {\normalfont\small $^1$Stanford University \quad $^2$University of Maryland, College Park \quad $^3$Harvard University} \\[1.2em]
  {\normalfont\small \hypersetup{pdfborder={0 0 0}}\textcolor{hotpink}{\url{https://masked-visual-actions.github.io}}}
}
\begin{document}

\maketitle

\vspace{-0.6cm}
\begin{figure}[h]
    \centering
    \includegraphics[width=\linewidth]{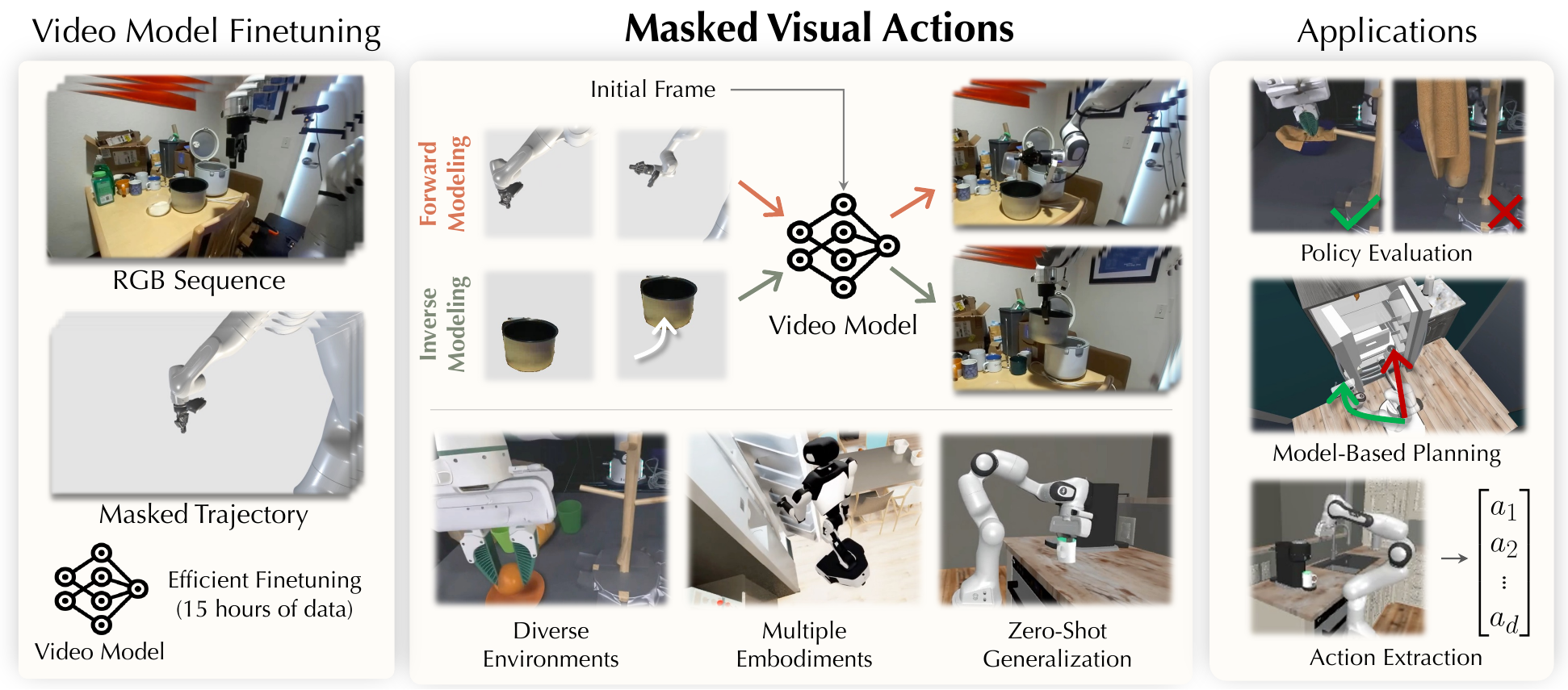}
    \vspace{-1.4em}
    \caption{\small \textbf{Masked Visual Actions.} We finetune a video model to condition on masked trajectories of robots, representing robot actions as pixel-space masked motions. Efficiently finetuned on only 15 hours of data, a single checkpoint of the model can act as an action-conditioned forward model to simulate robotic interactions with diverse and unseen embodiments. By conditioning it on object motion, it can also act as an inverse model that synthesizes the robot motion needed to achieve the desired outcome. We showcase the efficacy of our model for policy evaluation, model-based-planning, and action extraction where the video model acts as a policy.}
    \label{fig:teaser}
\end{figure}
\begin{abstract}
Video models absorb rich priors over how the visual world moves, interacts, and responds to contact, making them promising substrates for robotic world modeling.
The central challenge is how to communicate action to such models in a form aligned with the visual space in which they learned these interaction priors, yet still grounded in physical manipulation.
We introduce \textbf{Masked Visual Actions}, a pixel-space control interface that expresses action as
a partially revealed trajectory of an arbitrary entity in a video.
Revealing robot motion makes the model act as a forward dynamics model that predicts the scene's response to low-level robot actions, while revealing desired object motion makes the same model recover robot behavior consistent with that outcome.
Finetuned with only 15 hours of masked examples from real videos and simulation, a single checkpoint achieves strong visual fidelity and controllability across diverse scenes and multiple embodiments.
In downstream manipulation settings, the model produces imagined rollouts whose outcomes correlate with real-world execution for policy evaluation, improves decision making by ranking candidate futures in model-based planning, and supports inverse modeling by synthesizing robot motion from desired object motion.
\end{abstract}

\section{Introduction}
Purposeful interaction requires connecting an agent's actions to their effects in the world, and vice versa.
In sensorimotor control, skilled behavior is often described as coupling a forward model that anticipates the sensory consequences of movement with an inverse model that recovers the movement needed to realize a desired state~\citep{wolpert1995internal,wolpert1998internal,wolpert2001motor}.
Robotic world models should similarly support both directions of reasoning in a single predictive framework.

Recent advances in video models offer a promising route to this ambition. 
Trained on large-scale observation, they accumulate remarkably broad priors over motion, contact, persistence, deformation, and change, far beyond what can usually be distilled from robot data alone. 
However, most still remain passive observers rather than tools for intervention. 
Existing models condition generation on text~\citep{wan2025wan,nvidia2025cosmos}, tracks~\citep{geng2024motionprompting,chu2025wanmove,shin2026motionstream,singer2026timetomove,FangqiIRASim2024}, forces~\citep{gillman2025forceprompting,gillman2026goalforce,liu2026realwonder}, keypoints~\citep{xie2026generatedrealityhumancentricworld,wang2025vap}, or motor commands~\citep{gao2026dreamdojo,guo2026ctrlworld}, signals that are often sparse, embodiment-specific, or misaligned with the model's pre-trained visual experience.
What remains missing is an action representation expressed directly in the visual space where pretrained video models learned their interaction priors.
Once action is expressed \emph{visually}, the same model can complete different parts of an interaction depending on the trajectory revealed: revealing robot motion prompts a scene response, while revealing object motion prompts robot behavior.

To realize this vision, we introduce \textbf{Masked Visual Actions}, 
a method that recasts action as a visual primitive directly within the pre-trained model's native representation.
We finetune a pre-trained video model~\citep{wan2025wan} to ingest actions as partially revealed spatiotemporal patterns in pixel space---a masked trajectory of an entity in the scene. 
When the revealed entity is the robot, the model predicts the scene's response and acts as a forward dynamics model; when the revealed entity is instead an object or desired object motion, the same model acts as an inverse model to recover robot behavior consistent with that outcome.
In this view, active and passive roles are not properties of separate architectures, but different queries to the same interaction prior. While conceptually simple, this interface is pixel-aligned, embodiment-agnostic, native to video, and efficient to inject into a pretrained model through lightweight adaptation.

In addition to superior visual fidelity compared to prior work, we validate our framework across three applications in robot manipulation, both in simulation and in the real world: policy evaluation, model-based planning, and inverse modeling, where the video model is used as part of a robot policy.
All experiments use a \emph{single} checkpoint of the model finetuned on as few as 15 hours of robot interaction data.
In policy evaluation, the model's imagined rollouts exhibit consistent correlation with real-world outcomes, so that simulated performance serves as a useful proxy for actual execution.
In model-based planning, these same predictive capabilities are used to simulate the effects of different action trajectories and select the best one for execution, leading to consistent gains across diverse tasks and policy architectures.
The same checkpoint can also be used in reverse: given desired object motion, it synthesizes robot motion that achieves the goal, and a learned inverse dynamics model extracts the resulting actions.

We summarize our contributions as follows: 
\textbf{(1)} we introduce Masked Visual Actions, a pixel-space control interface for pretrained video models, together with an efficient adaptation recipe based on masked examples from real and simulated data; 
\textbf{(2)} we show that forward and inverse robot world-modeling problems can be cast as complementary conditional prediction problems of the same video model, obtained by revealing different entities in the scene; and 
\textbf{(3)} we validate this framework in both simulation and the real world across three applications in robot manipulation: policy evaluation, model-based planning, and inverse modeling.

\section{Related Work}
\begin{figure}[th]
    \centering
    \includegraphics[width=\linewidth]{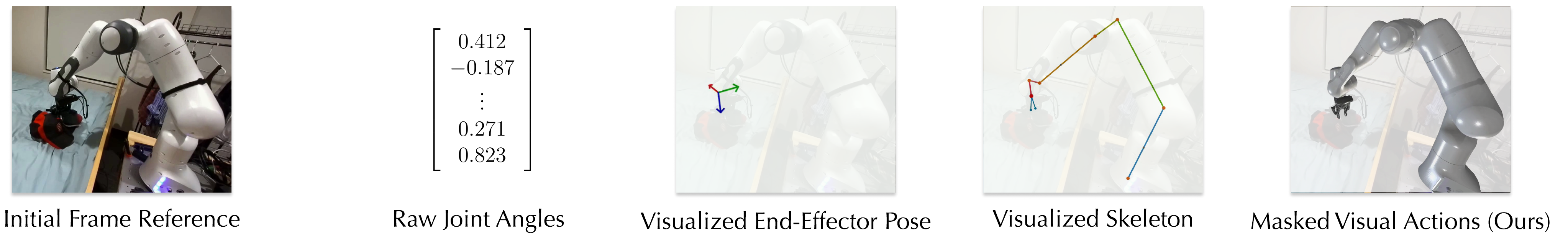}
    \caption{\small \textbf{Comparing action representations for learning.}
Low-dimensional robot actions are compact, but embodiment-specific and not aligned with the image observations used by video models. End-effector poses or robot skeletons are more visual, but remain sparse and require the model to infer geometry, contact, and interaction effects. Our masked visual actions provide dense, image-aligned conditioning, making robot motion and action directly visible, yielding a more learnable representation across embodiments and object interaction.}
    \label{fig:conditioning_comparison}
\end{figure}
\paragraph{Controllable video generation as a robotic interface.}
Video models become simulators once a control signal is added: physical forces \citep{gillman2025forceprompting,gillman2026goalforce}, warped flow \citep{burgert2025gowiththeflow}, hand poses \citep{xie2026generatedrealityhumancentricworld, kim2026dwm, goswami2025worlddexwm}, point or trajectory tracks \citep{shin2026motionstream,geng2024motionprompting,chu2025wanmove,singer2026timetomove}, and goal images \citep{huang2025vid}. 
None of these signals is dense, pixel-aligned, or shareable across embodiments.
Closer to us, visual prompting via inpainting~\citep{bar2022visual} and world modeling as conditional inference~\cite{kotar2025world} show that varying which region of an image is provided enables a generic task parametrization in visual domains; we extend this idea to robotic world modeling by using masked frames as the control interface.

\paragraph{Pixel-grounded action conditioning for robot world models.}
Most robotic video world models communicate actions through embodiment-specific channels: end-effector poses \citep{yang2023learning, FangqiIRASim2024,guo2026ctrlworld, qi2026inference}, joint vectors \citep{gao2026dreamdojo,guo2026ctrlworld}, or skeletons \citep{wang2025vap, veorobotics2025}. A growing line replaces them with pixel-grounded signals. BridgeV2W~\citep{chen2026bridgev2w} and Kinema4D~\citep{xu2026kinema4d} render the robot through its URDF and inject the resulting masks or pointmaps via ControlNet; Action Images~\citep{zhen2026action} encodes 7-DoF actions as multi-view Gaussian heatmaps; ORV~\citep{yang2025orv} conditions on 4D occupancy; Mask2IV~\citep{li2025mask} conditions on predicted mask trajectories; Mask World Model~\citep{lou2026mask} predicts semantic masks as the output. 
A complementary line treats masks as a data-editing tool: Shadow~\citep{lepert2025shadow, chen2025tool}, Phantom~\citep{lepert2025phantom}, Masquerade~\citep{lepert2025masquerade}, and EmbodiSwap~\citep{dessalene2025embodiswap} composite or render the robot onto human videos for cross-embodiment policy transfer. 
All of these works treat the robot as the active entity during training and run forward only. We expose the same masking interface to \emph{any} subset of entities, so one model serves as a forward, inverse, or unconditional generator without retraining.

\paragraph{Unified video-action models and downstream uses.}
UVA~\citep{li2025unified}, UWM~\citep{zhu2025unified}, AIM~\citep{fan2026aim}, X-WAM~\citep{guo2026unified}, and MotuBrain~\citep{team2026motubrain} unify forward dynamics, inverse dynamics, policy, and video generation in one model by masking modality channels (action vector vs.\ video) or by manipulating diffusion timesteps; large platforms such as Cosmos~\citep{nvidia2025cosmos}, Genie Envisioner~\citep{liao2025genie}, and DreamGen~\citep{jang2025dreamgen} package similar capabilities at foundation-model scale. Because the masking is over modality channels, actions remain low-dimensional vectors and the unification does not transfer across embodiments. 
Our masking is \emph{spatial}: active and passive entities live on the same pixel canvas, so the same forward/inverse switch also bridges the embodiment gap. The forward direction is exercised for policy evaluation \citep{zhang2026realpolicyeval,veorobotics2025,wang2026interactive}, policy improvement \citep{guo2026ctrlworld}, planning \citep{du2023learning, du2023video, zhen2025tesseract,  huang2026pointworld,chen2025largevideoplanner, rhoda2026dva}, and direct video-as-policy \citep{kim2026cosmos,hu2024video}; the inverse direction extracts robot motion through point tracking \citep{bharadhwaj2024track,karaev2024cotracker}, object flow \citep{ko2023learning, li2025novaflow,zhi2025dflowaction}, predicted object pose \citep{su2024motion}, or learned IDMs \citep{du2023learning, du2023video, wang2026eva,zhang2026veo,chen2025bimanual, pai2025mimic}. Critically, our inverse pipeline reuses the same backbone as the forward simulator, while prior work trains a separate IDM head or flow predictor.

\section{Masked Visual Actions}
Video generation models can be used to show how an initial scene evolves over time~\cite{wan2025wan,videoworldsimulators2024}, as they can model rich scene dynamics and object interactions. The video model captures the distribution $p(V)$ over videos $V \in \mathbb{R}^{T \times H \times W \times 3}$ depicting a scene. 
We view a scene $\mathcal{S}$ as a set of entities $e_1, e_2, ..., e_n$, and the video model generates a sequence of frames depicting how the entities interact over time. In the output video, each $e_i$ has a spatiotemporal trajectory, and we abuse notation slightly and write $e_i$ for both the entity and the spatiotemporal region of pixels it occupies.
Aligned with recent works built on structured masking and condition inference~\cite{kotar2025world,bear2023unifying,venkatesh2024understanding,loong2026zero}, the video model implicitly captures the joint distribution over all entity trajectories
%
\begin{equation}
p(V) = p(e_1, e_2, \ldots, e_n),
\end{equation}
including the interactions among them. Conditioning on a subset $\mathcal{S} \subseteq \{1, \ldots, n\}$ of entities yields the conditional distribution
\begin{equation}
p\bigl(\{e_i\}_{i \notin \mathcal{S}} \,\big|\, \{e_j\}_{j \in \mathcal{S}},\, I_0\bigr),
\end{equation}
where $I_0$ is a reference image of the initial scene. By varying $\mathcal{S}$, the same model answers different questions about the same scene.

We realize this conditioning by masking. Let $M \in \{0,1\}^{T \times H \times W}$ be a binary mask indicating which spatiotemporal pixels are revealed to the model ($M_{t,h,w} = 1$) versus predicted ($M_{t,h,w} = 0$). For a chosen conditioning set $\mathcal{S}$, the mask is the union of the pixel regions occupied by the conditioned entities,
\begin{equation}
M(\mathcal{S}) = \bigcup_{i \in \mathcal{S}} e_i,
\end{equation}
and the model receives as input the masked video $M \odot V$ together with the reference image $I_0$. Training proceeds by sampling $M$ from a distribution over masks and learning the conditional $p_\theta(V \mid M \odot V, I_0)$. 
We draw inspiration from masked modeling in language \citep{devlin-etal-2019-bert} and masked-image prompting \citep{bar2022visual}, where varying the masked input enables diverse applications with the same model. 

\begin{figure}[t]
    \centering
    \includegraphics[width=\linewidth]{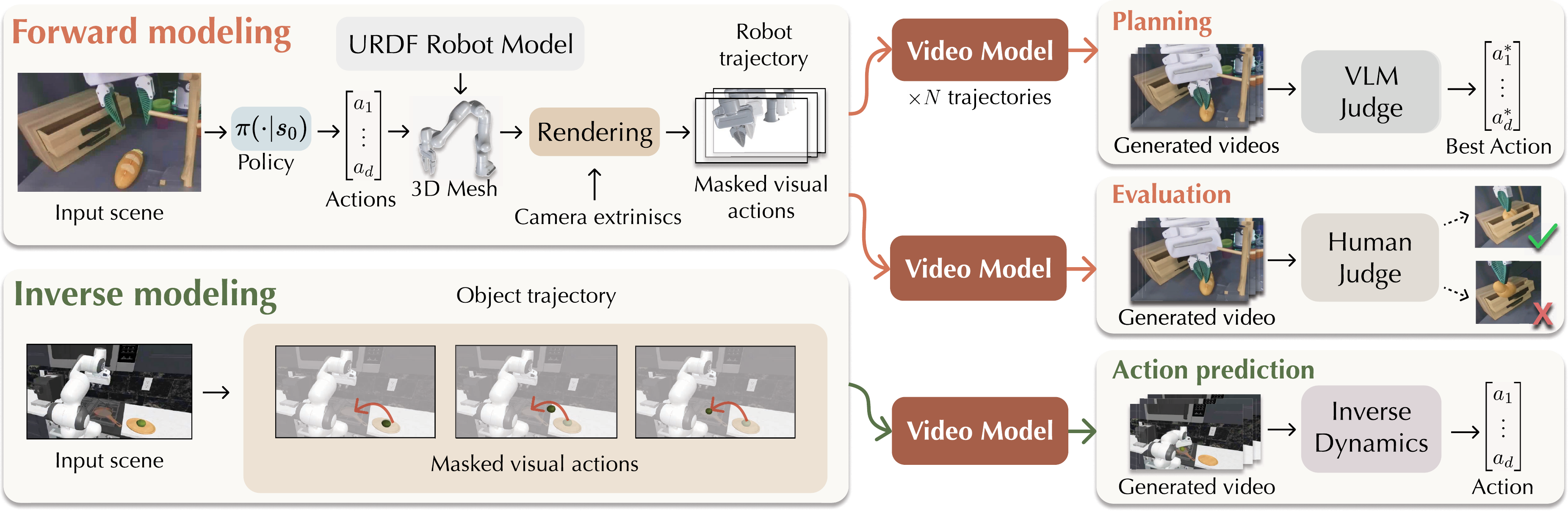}
    \vspace{-1.5em}
    \caption{\small \textbf{Applications.} The masked visual actions allow using the video model as a forward model, conditioned on robot actions, or as an inverse model that predicts the robot motion that satisfies the object trajectory. The forward model can be used for planning and choosing the best trajectory sampled from a  policy, or policy evaluation. On the other hand, the inverse modeling can be combined with an inverse dynamics model to estimate robot actions from the generated video.}
    \label{fig:applications}
    \vspace{-1.5em}
\end{figure}

In robotics, it is convenient to partition entities into two roles, as illustrated in Figure~\ref{fig:applications}. We call entity $e_i$ \emph{active} if it acts on the scene through its own agency, such as a robot arm or a human, and \emph{passive} if its motion arises from interaction with an active entity, such as a manipulated object. Let $\mathcal{A} \subseteq \{1, \ldots, n\}$ index the active entities and $\mathcal{P} = \{1, \ldots, n\} \setminus \mathcal{A}$ the passive ones. This partition surfaces two natural ways to use the model.

\paragraph{Forward model.} Setting $\mathcal{S} = \mathcal{A}$, we condition on the active entities and predict the passive ones,
\begin{equation}
p\bigl(\{e_i\}_{i \in \mathcal{P}} \,\big|\, \{e_j\}_{j \in \mathcal{A}},\, I_0\bigr).
\end{equation}
This corresponds to the standard action-conditioned dynamics modeling, in which a robot's motion is provided, and the model simulates its effect on the scene. Unlike prior work that conditions on low-dimensional action commands \citep{gao2026dreamdojo,guo2026ctrlworld}, the active conditioning here is supplied as masked videos that are agnostic to embodiments.

\paragraph{Inverse model.} Setting $\mathcal{S} = \mathcal{P}$, we condition on the passive entities and predict the active ones,
\begin{equation}
p\bigl(\{e_i\}_{i \in \mathcal{A}} \,\big|\, \{e_j\}_{j \in \mathcal{P}},\, I_0\bigr).
\end{equation}
This direction has no analog in conventional action-conditioned world models: the user specifies a desired outcome in the world, and the video model recovers the agent behavior consistent with it.

The active/passive distinction is a convenient way to describe how we use the model rather than a property of the model itself. The model is trained on masked video completion without any explicit notion of agency, and at inference, any subset $\mathcal{S}$ can be chosen. In fact, we trained our model only on masks depicting active robotic entities, yet it generalizes to queries conditioned on passive entities in a \emph{zero-shot} manner. As observed in our empirical evaluations, this behavior is unique to our masked visual action conditioning, as conditioning the video model on sparser signals, such as low-level action commands or visualized skeletons, cannot achieve this level of generalization. 

\section{Method}



\begin{figure}[th]
    \centering
    \includegraphics[width=\linewidth]{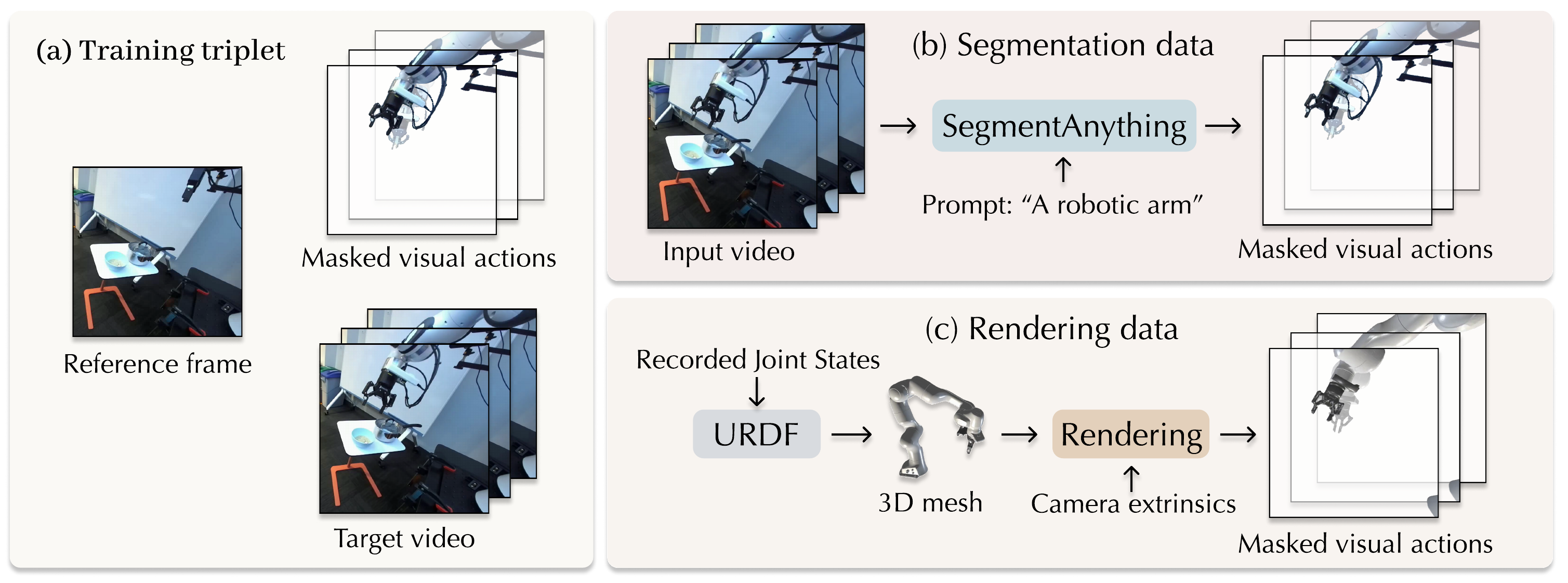}
    \vspace{-1.5em}
    \caption{\small \textbf{Dataset construction.} (a) to train our model, we need a reference frame of the initial scene, and the masked visual actions, and train it to reproduce a realistic video of the robot executing the input actions. (b) We use segmentation-based approach by segmenting the robot arm from robotics datasets as the masked visual actions. (c) However, to allow the user to provide arbitrary action trajectory at inference, the model needs to also accept simulated mesh visualization of those actions. As a result, we also include rendering-based dataset. Given that DROID also contains the robot state at each timestep, we render the robot URDF that matches the original video to construct masked visual actions.}
    \label{fig:dataset}
    \vspace{-1.2em}
\end{figure}
\subsection{Dataset construction}
We construct the masked modeling dataset by combining real-world videos from DROID~\citep{khazatsky2024droid} and simulation data from Robocasa~\citep{robocasa365}. We use both success and failure trajectories from both datasets. 
We follow two approaches to construct masked conditioning for each video, based on video segmentation and rendering the robot state, as outlined below.
\paragraph{Segmentation based dataset}
Given any video, we can segment any entity in the scene using SegmentAnything~\citep {carion2026sam3}, without the need for camera calibration or even explicitly knowing which robot is shown in the video. 
We use videos from DROID, and use the prompt ``A robotic arm'' for segmentation to isolate the robot. 
Using segmentation data enables the model to effectively learn to \textit{inpaint} missing regions and to model the joint distribution over all entities in the scene. While a segmentation-based approach is highly general, it suffers from two major limitations: First, it is challenging for the user to provide an exact segmentation mask of entities at test time. Second, any occluded regions in the robot would implicitly leak information about the scene dynamics from the original video. To mitigate those limitations, we also explore the approach that explicitly renders robots from their recorded state.
\paragraph{Rendering based dataset}
Instead of relying on segmentation, we can also align a robot mesh with the input video and use that as the masked conditioning. By rendering the robot mesh, we can visualize arbitrary action trajectories during inference and then use them as masked conditioning for the video model.
To construct a rendering-based dataset, we require the robot state corresponding to the input video and the camera calibration. 
We use the DROID dataset and follow the protocols from PointWorld~\cite{huang2026pointworld} to refine the camera calibration to accurately align the robot URDF with the input trajectories. 
In the Robocasa simulation, we render only the robot, excluding the rest of the scene, to generate the masked conditioning. 
To allow the model to see the full robot without self-occlusion, we render the robot only with translucent rendering and set the gripper fingers to bright red so the video model can easily observe the actions. 
Note that the rendering approach requires known camera calibration, and is limited to rendering the robot only as opposed to arbitrarily enabling masking any entity in the scene. As a result, we believe that the segmentation and rendering-based approaches are complementary.


\subsection{Model implementation and training}
We use Wan-Fun-Control 2.2 14B~\cite{wan2025wan} as the base model. We encode the masked conditioning video using the same autoencoder as the video model and use concatenation as the conditioning mechanism. Concatenation is appropriate as the conditioning signal is spatially aligned with the desired output video. For the missing region from the masked conditioning, we set it to a uniform gray background.
Instead of finetuning the entire model, we use LoRA finetuning with rank 256, and a batch size of 4 using 8 NVIDIA H200 GPUs. 
We train the model for approximately 10,000 steps over 4 days. \textit{For reproducibility, we will release our code, data, and model weights.}

\begin{wrapfigure}{r}{0.5\textwidth}
    \centering
    \vspace{-1.5em}
    \includegraphics[width=\linewidth]{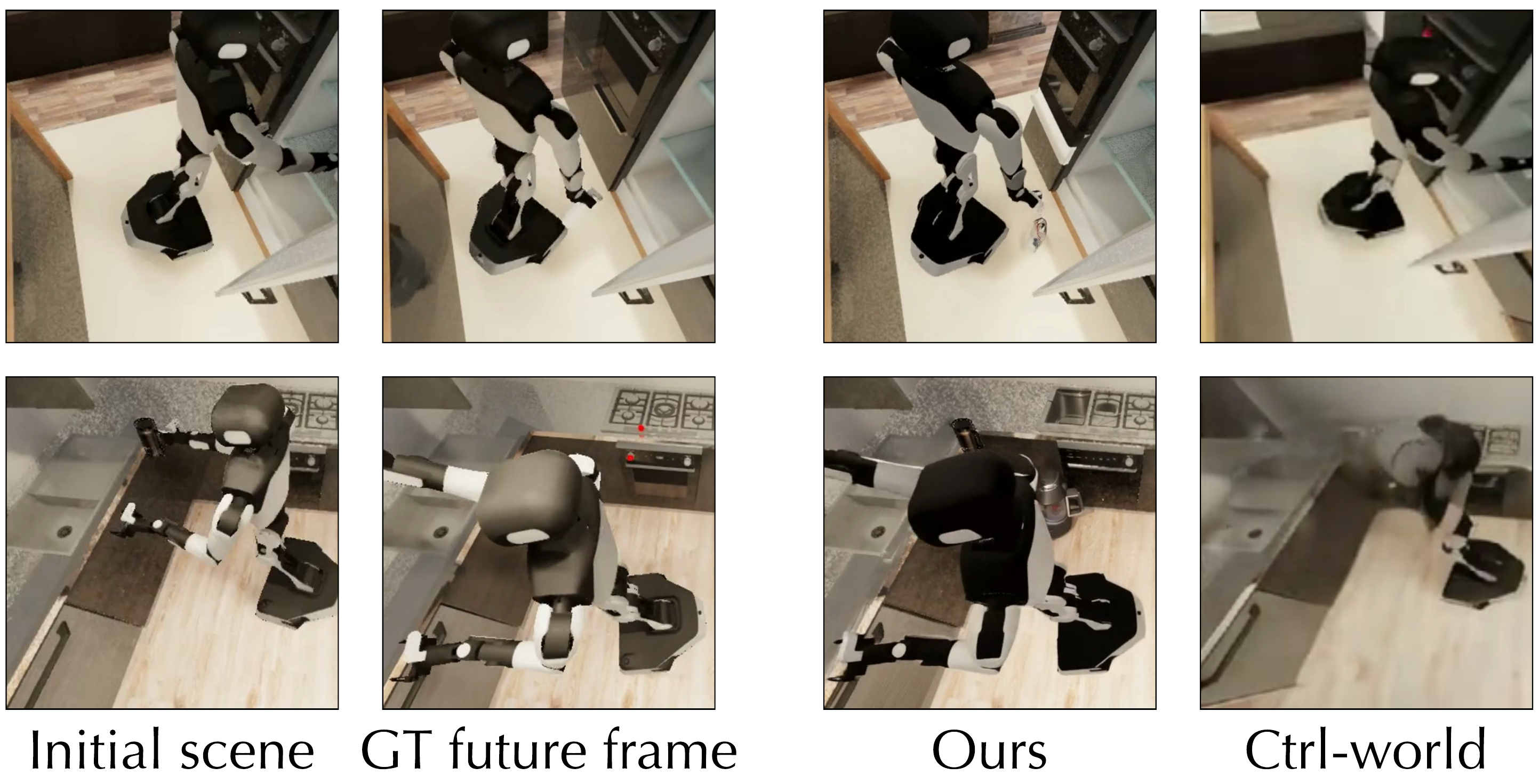}
    \vspace{-1.5em}
    \caption{\small \textbf{Generalization to unseen embodiment.} While using the raw action state such as in Ctrl-world~\cite{guo2026ctrlworld} can work well within the training domain, it collapses on unseen embodiments. However, our method can generalize well to unseen embodiments.}
    \label{fig:qual_behavior_compariosn}
    \vspace{-2.5em}
\end{wrapfigure}

\section{Experiments}
We start by evaluating Masked Visual Actions as a control signal for world modeling. 
We evaluate visual fidelity and controllability against prior work and highlight generalization to embodiments unseen during training. 
Afterward, we evaluate diverse robotic applications of our video model. 
In particular, we show how it can be used for planning by evaluating sampled trajectories, for policy evaluation, and for using the video model as the policy itself through inverse modeling.
\textit{Please refer to the project webpage for video results.}

\subsection{Controllable video generation}
\begin{figure}[t]
    \centering
    \includegraphics[width=\linewidth]{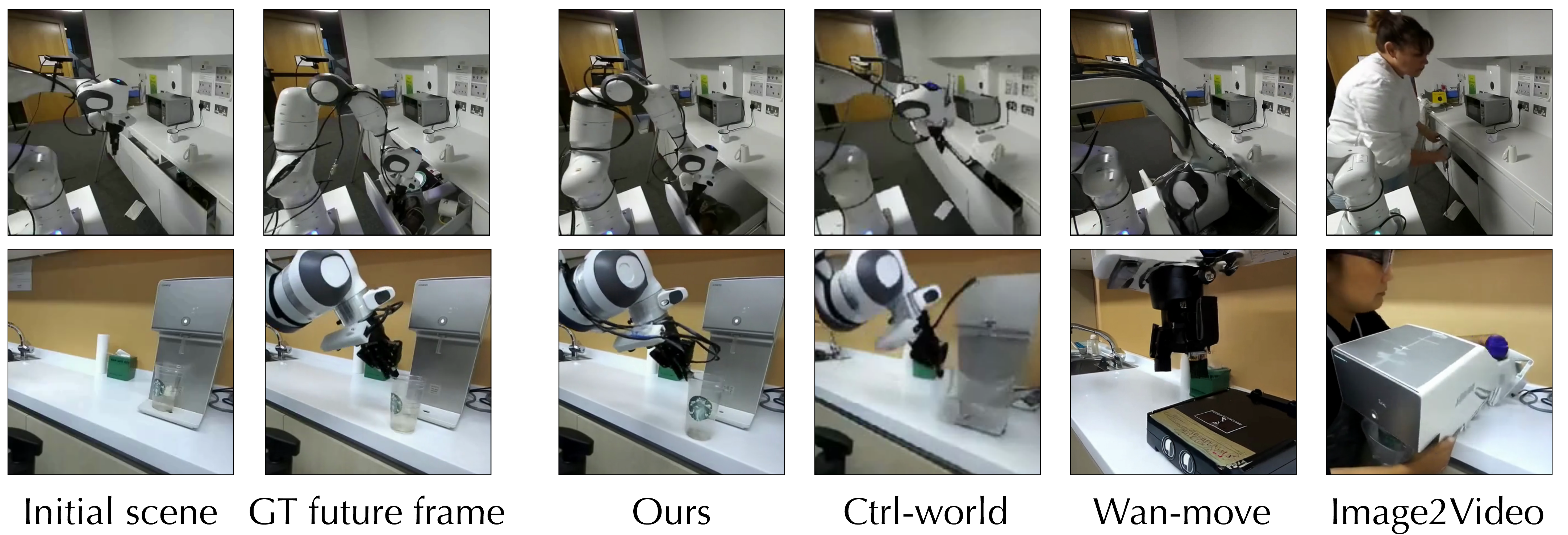}
    \caption{\small \textbf{Comparing baselines on DROID.} Using image-to-video~\cite{wan2025wan} or even trajectory conditioned video generation~\cite{chu2025wanmove} with GT tracks fails to execute the robot motion or preserve the input scene. On the other hand, our model can competitively match and outperform models that take the raw robot actions~\cite{guo2026ctrlworld} while maintaining generalization.}    \label{fig:qual_droid_compariosn}
    \end{figure}
\begin{table}[t]
\centering
\caption{\small \textbf{Baseline comparison on diverse embodiments.} We evaluate our method against Ctrl-World~\cite{guo2026ctrlworld} on DROID as a seen robotic embodiment, as well as BEHAVIOR, which uses a bimanual robotic embodiment that's unseen for all methods. Our model outperform the baseline on both datasets, and we include image-to-video and trajectory conditioned video models as a reference.}
\vspace{-0.2em}
\label{tab:video_results}
\small
\setlength{\tabcolsep}{5pt}
\begin{tabular}{l ccc ccc}
\toprule
 & \multicolumn{3}{c}{DROID} & \multicolumn{3}{c}{BEHAVIOR} \\
\cmidrule(lr){2-4} \cmidrule(lr){5-7}
Method & LPIPS $\downarrow$ & SSIM $\uparrow$ & PSNR $\uparrow$ & LPIPS $\downarrow$ & SSIM $\uparrow$ & PSNR $\uparrow$ \\
\midrule
Image-to-video \citep{wan2025wan} & 0.521 & 0.548 & 12.42 & 0.602 & 0.457 & 10.22 \\
Wan-move~\citep{chu2025wanmove} & 0.534 & 0.562 & 12.99 & 0.312 & 0.756 & 13.17 \\
Ctrl-World \citep{guo2026ctrlworld}  & 0.362 & 0.708 & 18.15 & 0.196 & 0.837 & 18.39 \\
\midrule
Masked Visual Actions (Ours) & \textbf{0.0945} & \textbf{0.887} & \textbf{23.74} & \textbf{0.123} & \textbf{0.843} & \textbf{22.90} \\
\bottomrule
\vspace{-2.0em}
\end{tabular}
\end{table}

Video generation traditionally conditioned on text is expressive, but underspecified.
Conditioning on motion tracks preserves the generality while allowing us to condition on motion.
On the other hand, conditioning on the action space for a specific robotic embodiment provides additional precision, but at the cost of generality.
Through Masked Visual Actions, we aim to preserve the generality of the video model by visually conditioning the video model on the robotic actions, and setting the role of the video model to answer: \textit{given this visual masked action, how would the rest of the scene look like?}
As a baseline for using the robot's raw end-effector state as input, we use Ctrl-world~\cite{guo2026ctrlworld}, a recent SoTA method. 
For track conditioning, we use Wan-move~\cite{chu2025wanmove}, conditioning it on ground-truth tracks computed from the robot mesh. Additionally, we include Wan2.2 14B image-to-video as a reference. 

In Fig.~\ref{fig:qual_droid_compariosn}, we highlight that both Masked Visual Actions and Ctrl-world accurately follow the robot actions on our held out scenes from DROID\footnote{Ctrl-World was trained on the entirety of DROID, so it has seen our held out scenes during training.}
On the other hand, both Wan-Move and Wan I2V completely collapse and transform the input scene. However, unlike our method, conditioning on the raw robot state cannot generalize to unseen embodiments~\cite{RoboPanoptes25,li2024behavior1k}. We use data from BEHAVIOR~\cite{li2024behavior1k}, which uses a bimanual robot, R1-Pro, to evaluate generalization on unseen embodiments. In Fig.~\ref{fig:qual_behavior_compariosn}, we demonstrate that Ctrl-world simply outputs static or corrupted videos for unseen embodiments, while our model can gracefully handle the unseen embodiment. 
We quantitatively evaluate performance on generated videos across both DROID and BEHAVIOR in Table~\ref{tab:video_results} and show that our method outperforms the baselines.

\begin{figure}[t]
    \centering
    \includegraphics[width=\linewidth]{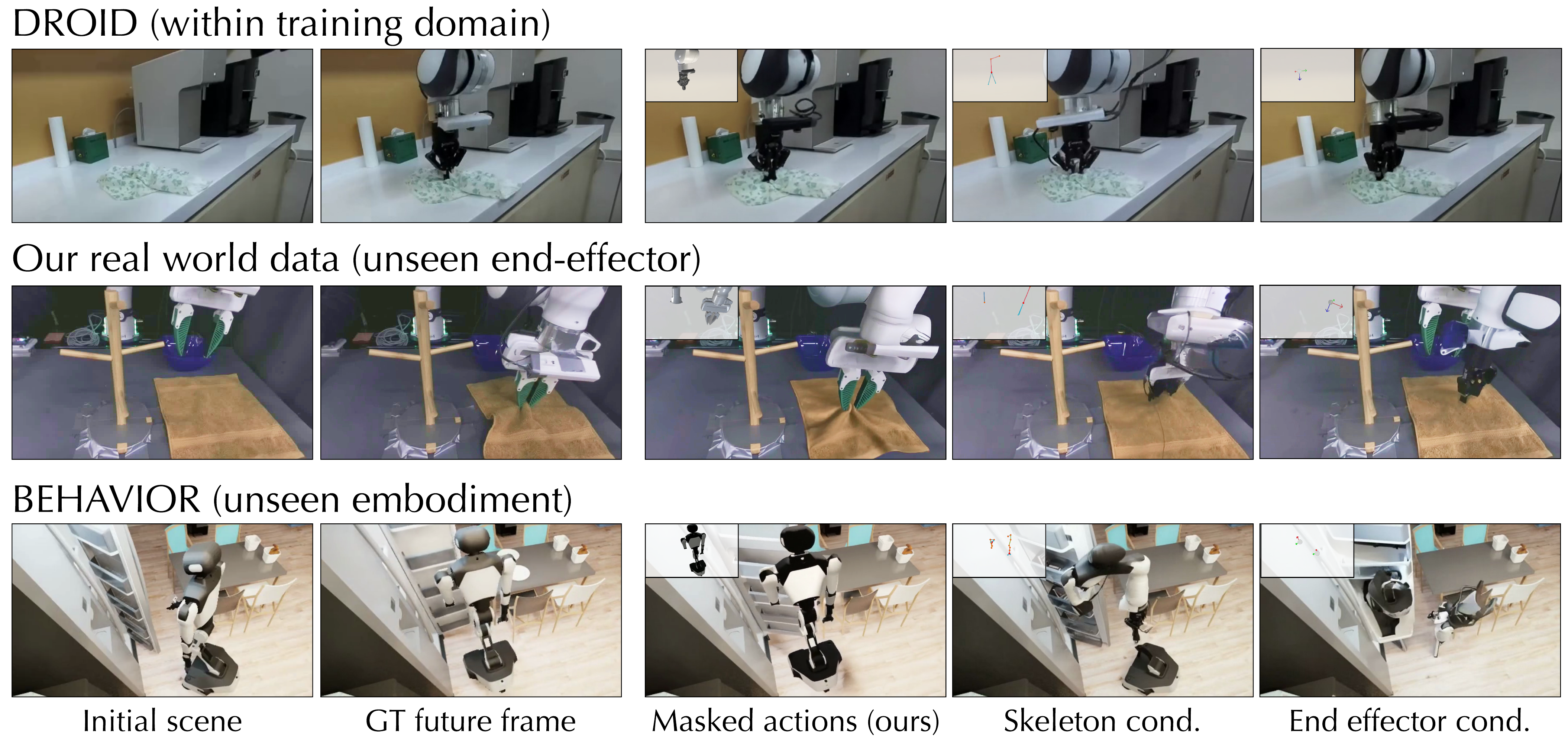}
    \vspace{-1.6em}
    \caption{\small \textbf{Comparing action conditioning.} Training a video model on different conditioning signals on DROID such as masked visual actions, end effector visualization, or skeleton all work well within the training domain. However, when going beyond the training distribution, such as using a custom end-effector, the models trained on skeleton and end effector position would hallucinate the robot seen in training or transform the robot to match the training. Furthermore, on unseen embodiments such as bimanual robots in BEHAVIOR, using masked actions generalizes gracefully, while other conditioning signals transform and disfigure the robot.}
    \label{fig:ablating_action_cond}
\end{figure}
\begin{table}[t]
\centering
\caption{\small \textbf{Ablation on visual conditioning signal.} When using sparse conditioning signal, the performance on held out data from the training distribution on DROID is similar to using masked visual actions. However, when using the same robot with unseen gripper (such as on our real world data), or on a robot from unseen embodiment in BEHAVIOR, the gap increases significantly between our masked actions and the other conditioning methods.}
\label{tab:action_ablation}
\small
\setlength{\tabcolsep}{2pt}
\begin{tabular}{l ccc ccc ccc}
\toprule
 & \multicolumn{3}{c}{DROID} & \multicolumn{3}{c}{Real world}  & \multicolumn{3}{c}{BEHAVIOR} \\
\cmidrule(lr){2-4} \cmidrule(lr){5-7} \cmidrule(lr){8-10}
Method & LPIPS $\downarrow$ & SSIM $\uparrow$ & PSNR $\uparrow$ & LPIPS $\downarrow$ & SSIM $\uparrow$ & PSNR $\uparrow$ & LPIPS $\downarrow$ & SSIM $\uparrow$ & PSNR $\uparrow$ \\
\midrule
End-effector vis. & 0.107 & 0.878 & 22.64 & 0.183 & 0.858 & 20.32 & 0.171 & 0.815 & 19.23 \\
Skeleton vis. & 0.106 & 0.878 & 22.74 & 0.169 & \textbf{0.866} & 21.02 & 0.162 & 0.824 & 19.58 \\
\midrule
Masked Visual Actions & \textbf{0.0945} & \textbf{0.887} & \textbf{23.74} &  \textbf{0.148} & 0.864 & \textbf{22.79} & \textbf{0.123} & \textbf{0.843} & \textbf{22.90} \\
\bottomrule
\vspace{-1.0em}
\end{tabular}
\end{table}

\paragraph{Comparing the choice of visual actions}
Conditioning on visual actions allows diverse ways to represent the action. 
We compare against visualizing the end-effector pose, inspired by IRASim~\cite{FangqiIRASim2024}, and the robot skeleton, adopted in VAP~\cite{wang2025vap}. 
We train the same base model used for our method, and use the same training dataset from DROID to train the baselines. 
While we expect the varying action conditioning to perform similarly on the same domain as the training set, sparse conditioning signals require the model to explicitly learn the correspondence between the sparse action and the target video. 
However, by conditioning on masked visual actions, the model simply needs to model the interaction between the masked input and the rest of the scene. In Fig.~\ref{fig:ablating_action_cond}, we show that on DROID, all the variants of our model perform similarly. However, on real-world data we captured using a similar robot to the one used in DROID, the Franka Emika Panda, but with a custom 3D-printed end-effector, we find that using a sparse conditioning signal suffers significantly. 
In particular, when conditioning on the robot skeleton, the video model would transform the robot to match the embodiment seen during training. When using the end-effector visualization as input, the model would simply introduce another robot into the scene that matches the training data. 
To further deviate from the training setting, we test the models on the R1 Pro in BEHAVIOR~\cite{li2024behavior1k}. Given that R1 Pro has two end effectors, we adapt the baseline visualizations to show two end effectors and the skeleton poses of each. We find that conditioning on the end-effector or skeleton visualization completely collapses and distorts the robot. However, when using masked visual actions, the model can gracefully generalize and simulate the physical interaction of the robot opening the fridge.  
In Table~\ref {tab:action_ablation}, we quantitatively evaluate the video generation performance for masked visual actions, and the sparser conditioning mechanisms of end effector pose visualization and robot skeleton.



\subsection{Robotics applications}
We highlight multiple applications of our unified world model in robotics. We use our model as a forward model to simulate robot actions and demonstrate its use for planning and policy evaluation. We also use our model as an inverse model: given the desired object motion as a masked visual action, we generate a video of the robot performing the desired object manipulation and extract the actions using a learned inverse dynamics model. Across these applications, we use Robocasa~\citep{robocasa365} as the simulation environment.

\paragraph{Planning}
\label{sec:planning}

\begin{figure}[t]
    \centering
    \includegraphics[width=\linewidth]{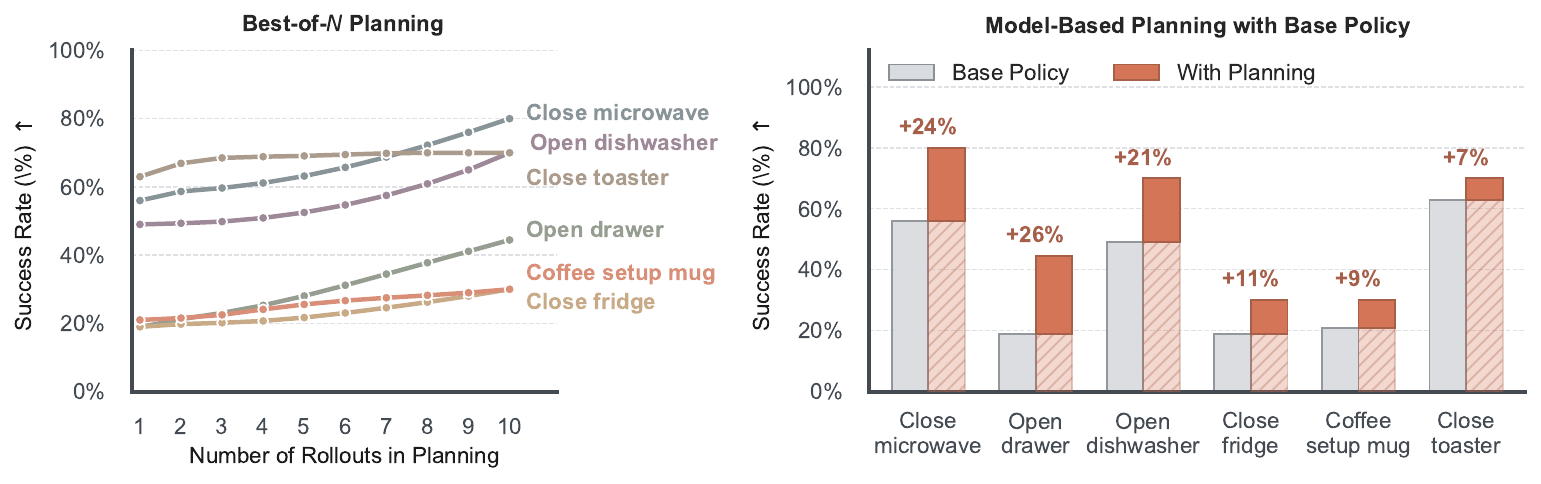}
    \vspace{-2.0em}
    \caption{\small \textbf{Application on Planning.} By rolling out multiple trajectories from a pretrained diffusion policy, we can evaluate each trajectory by simulating the actions with the video model, and then using a VLM judge to pick the best action trajectory. We observe consistent improvement in task success when using the video model to roll out and choose best action sequences, as well as the positive correlation with the number of samples evaluated at test time. This demonstrates the ability of the model to simulate counterfactuals given the same initial condition.}
    \label{fig:planning}
    \vspace{-1.5em}
\end{figure}
Given the same environment observation, multiple rollouts sampled from a stochastic policy may achieve varying levels of task progress. 
By rolling them out in an action-conditioned video model, one may evaluate the trajectories purely in imagination before executing them in the actual environment. 
In our experiments, we use Diffusion Policy~\cite{chi2023diffusionpolicy, chi2024diffusionpolicy} as the stochastic policy and Best-of-N as the simplest model-based planning algorithm.
After simulating the action candidates using the video model, we evaluate each rollout with Gemini 3.1 Pro to assess their relative task success, interaction fidelity, and physical realism. 
We evaluate on 10 scenes per task, with $N=10$. We include the detailed criteria in the appendix. After evaluating all the rollouts, we pick the best action sequence to execute. 
In Fig.~\ref{fig:planning}, we highlight the improvement in performance on diverse tasks and show how  success rate increases with the number of action samples. This approach can be viewed as a form of test-time scaling~\citep{muennighoff2025s1testtimescaling, ma2025scalingdiff}, leveraging additional compute to achieve higher performance. 
In our case, the policy and video model act as the generator, and the VLM critic acts as the verifier.


\paragraph{Policy evaluation}
We can also use our model to evaluate policy performance by comparing a policy's success rate within the video model to that within the ground-truth environment. 
Similarly to the section above, we use an open-loop diffusion policy across tasks and rollout 10 trajectories per scene. 
We simulate action trajectories using our model and manually evaluate each rollout as a success or failure based on predefined task rubrics. 
\begin{figure}[t]
    \centering
    \begin{minipage}[t]{0.45\textwidth}
        \centering
        \includegraphics[height=4.4cm]{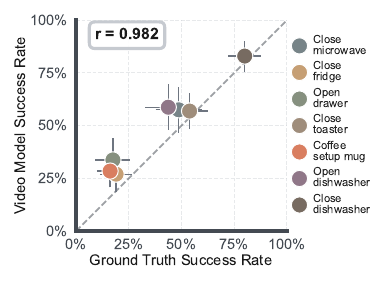}
        \captionof{figure}{\textbf{Robocasa policy evaluation.} Video-model rollouts consistently track ground-truth success rates across RoboCasa tasks.}
        \label{fig:policy_evaluation}
    \end{minipage}
    \hfill
     \begin{minipage}[t]{0.51\textwidth}
        \centering
        \includegraphics[height=4.4cm]{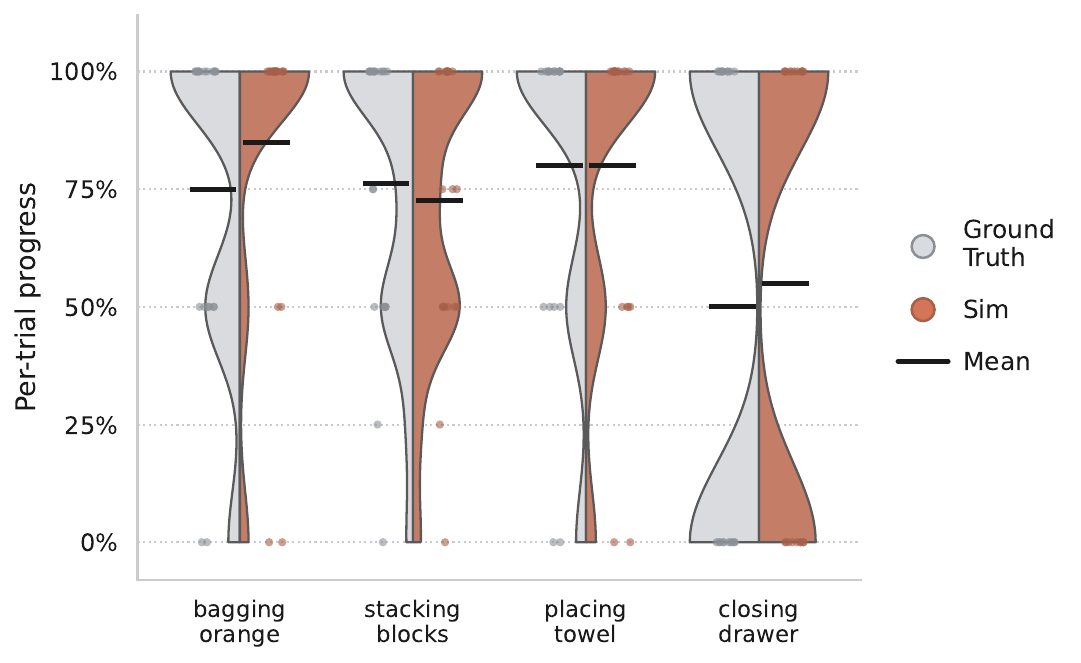}
        \captionof{figure}{\textbf{Real world policy evaluation.} Rolling out real-world demonstrations with our video model produces videos with success progress closely aligned with what is observed in the real world execution.}
        \label{fig:real_policy_evaluation}
    \end{minipage}
\end{figure}
The simulated rollouts are additionally evaluated by physical interaction realism (e.g., hallucinated task progress without contact is considered failure). In Fig.~\ref{fig:policy_evaluation}, we plot the success rate of each policy in GT environment against that evaluated within the video model, which exhibits a strong correlation with $r=0.982$.
\begin{wrapfigure}{r}{0.33\textwidth}
    \centering
    \vspace{-0.1em}
    \includegraphics[width=\linewidth]{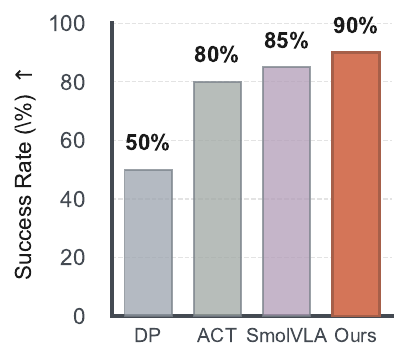}
    \vspace{-2.3em}
    \caption{\small \textbf{Action extraction.} Even without task-specific video-model training, inverse modeling recovers competitive robot behavior.}
    \label{fig:action_extraction}
    \vspace{-0.1em}
\end{wrapfigure}

However, we observe that the video model shows a positive bias towards task progress, as evidenced by consistently higher task success rates in its imagination. 
Beyond simulation, we evaluate our model in a real-world setup. For each of four tasks we collect 20 demonstrations, roll out each demonstration with the video model, and score both the real and simulated executions with a per-task rubric measuring partial task progress.
For each task, we collect 20 demonstrations, and set a rubric for evaluating the success progress for each demonstration.
Because each simulated rollout is paired with the demonstration it was generated from, we compare progress both in distribution and per trial. In Fig.~\ref{fig:real_policy_evaluation} we plot the per-trial progress distribution in the video model against the ground-truth distribution for each task. The two distributions closely match, but similar to simulation, it shows a positive bias towards task progress.
We include each task's rubric in the appendix and all generated videos and GT demonstrations in the project webpage. 
\paragraph{Action extraction}
Instead of providing the model with robot actions to execute, we can alternatively leverage Masked Visual Actions to formulate an inverse modeling problem: given desired object motion, prompting the video model to synthesize a video of a robot achieving that outcome.
Action extraction can then be cast as an inverse-dynamics problem: given the synthesized robot video, recover an executable low-level action sequence with a learned inverse-dynamics model.
We initially expected this setting to require explicit inverse-modeling finetuning. Instead, the video model trained only on forward examples already generalizes zero-shot to the inverse setting, likely because the conditioning signal is well-aligned with the model's learned representation.
We evaluate this pipeline on \textsc{CoffeeServeMug} in RoboCasa, where the robot must reach, grasp, transport, and place a mug from the coffee machine onto the table in a tightly constrained workspace.
We compare to standard imitation learning baselines, including Diffusion Policy~\citep{chi2023diffusionpolicy, chi2024diffusionpolicy}, ACT~\citep{zhao2023learning}, and SmolVLA~\citep{shukor2025smolvla}.
The inverse-dynamics model and all baselines are trained on 100 demonstrations, whereas the video model itself has not seen examples from this task.
Each method is evaluated with 20 trials, with success rates reported in Figure~\ref{fig:action_extraction}.
Our method achieves the highest success rate at $90\%$. This indicates that, although the video model has not been trained on an inverse modeling problem, it can be effectively prompted to extract robot behaviors from its rich interaction priors, while maintaining the competitiveness of modern imitation learning methods.
\section{Discussion and Conclusions}
By finetuning a pretrained video model on a small amount of Masked Visual Actions data, we efficiently leverage the prior of the video model to synthesize counterfactuals by conditioning on a subset of scene entities. 
Our model can simulate robot actions when conditioned on robotic embodiment visualization as an action-conditioned forward model, and when acting as an inverse model, where it synthesizes suitable robot motion to realistically manipulate the object.

\textbf{Limitations\;} It is worth noting that our model, similarly to existing generative models, learns the \emph{correlation} between object interaction rather than \emph{causal relationships}, which remains an open research question. 
Furthermore, our method is naturally limited by the base video model's capabilities, in terms of both inference speed and what it can express, as it re-purposes the model's prior rather than modifying its capabilities.

\textbf{Societal and broader impact\;} 
By enabling video-based policy evaluation, planning, and inverse modeling, our work could lower the cost of developing robotic systems and make robot learning more accessible. 
However, the same capabilities could also be used for unsafe or unauthorized robotic behaviors, highlighting the need for responsible use and deployment.


















\bibliographystyle{plainnat}
\bibliography{refs,references}

\newpage
\appendix

\setcounter{section}{0}
\setcounter{figure}{0}
\setcounter{table}{0}
\setcounter{equation}{0}

\renewcommand{\thesection}{\Alph{section}}
\renewcommand{\thefigure}{\Alph{section}\arabic{figure}}
\renewcommand{\thetable}{\Alph{section}\arabic{table}}
\renewcommand{\theequation}{\Alph{section}\arabic{equation}}

\section*{Appendix Overview}

This appendix provides detailed supplementary material supporting the main paper on video-based world models for robotic manipulation. It is organized into eight sections covering implementation details, evaluation protocols, and data collection procedures:

\begin{enumerate}
  \item \textbf{Project webpage}: References to video results and qualitative demonstrations of the model's performance on manipulation tasks.

  \item \textbf{Reproducibility}: Commitment to release complete code, model weights, and training data to enable reproduction of results.

  \item \textbf{Full quantitative results}: Comprehensive tables presenting quantitative metrics across all reconstruction experiments with standard error bars, demonstrating consistent improvements over baselines.

  \item \textbf{Additional training details}: Dataset composition and construction, including use of DROID (1,000 demonstrations) and Robocasa (4,000 examples), handling of failure cases, and conditioning strategy for the video model.

  \item \textbf{VLM Evaluation Protocol}: Automated evaluation methodology using Gemini 3.1 Pro Preview to assess generated robot videos. Details the Gemini system prompt used, task-agnostic failure detection (ghost contact, post-disengagement coasting, frame-jump glitches), structured JSON output format, rollout ranking via lexicographic ordering, and subset enumeration for success-rate curves. 

  \item \textbf{Policy and Inverse-Dynamics Training Details}: Training procedures for three policy settings—simulation policies for planning/evaluation in RoboCasa, inverse-dynamics model for action extraction from generated videos, and real-world policies for physical robot studies. Covers Diffusion Policy, ACT, and SmolVLA baselines.

  \item \textbf{Robot Data Collection}: Hardware setup, calibration procedures, and data representation. Describes Franka Panda manipulator with custom end-effector, dual ZED Mini 2 cameras (wrist-mounted and external), AprilTag-based hand-eye calibration, and HDF5 trajectory storage with camera intrinsics and extrinsics.

  \item \textbf{Real World Tasks Rubrics}: Description of the real-world tasks used for policy evaluation, and the scoring system used to determine the success progress.

  \item \textbf{Adapting Baselines to Unseen Embodiments}: Methodology for evaluating generalization to unseen robot morphologies using BEHAVIOR-1K benchmark. Describes adaptation strategies for the video model (direct URDF conditioning), Ctrl-World baseline (heuristic-based active arm identification), and skeleton/pose visualizations for bimanual robots.
\end{enumerate}

\section{Project webpage}
Please refer to the project webpage for video results at \hypersetup{pdfborder={0 0 0}}\textcolor{hotpink}{https://masked-visual-actions.github.io}

\section{Reproducibility}
To ensure full reproducibility, we release our code and model weights. 

\section{Full quantitative results}

 \begin{table}[t]
  \centering
  \caption{Reconstruction metrics across datasets (mean $\pm$ SEM; best per column in bold).}
  \label{tab:recon-all}
  \setlength{\tabcolsep}{3pt}
  \resizebox{\textwidth}{!}{%
  \begin{tabular}{l ccc ccc ccc}
  \toprule
  & \multicolumn{3}{c}{BEHAVIOR-1K ($n{=}50$)} & \multicolumn{3}{c}{DROID ($n{=}50$)} & \multicolumn{3}{c}{Real-World ($n{=}13$)} \\
  \cmidrule(lr){2-4} \cmidrule(lr){5-7} \cmidrule(lr){8-10}
  Method & PSNR$\uparrow$ & SSIM$\uparrow$ & LPIPS$\downarrow$ & PSNR$\uparrow$ & SSIM$\uparrow$ & LPIPS$\downarrow$ & PSNR$\uparrow$ & SSIM$\uparrow$ & LPIPS$\downarrow$ \\
  \midrule
  Ctrl-World & $18.39{\pm}0.54$ & $0.836{\pm}0.008$ & $0.196{\pm}0.011$ & $18.15{\pm}0.16$ & $0.708{\pm}0.008$ & $0.362{\pm}0.006$ & --- & --- & --- \\
  Wan-Move   & $13.17{\pm}0.30$ & $0.528{\pm}0.015$ & $0.454{\pm}0.016$ & $12.96{\pm}0.37$ & $0.558{\pm}0.017$ & $0.534{\pm}0.023$ & --- & --- & --- \\
  Wan2.2-I2V & $10.22{\pm}0.38$ & $0.457{\pm}0.023$ & $0.602{\pm}0.024$ & $12.42{\pm}0.34$ & $0.548{\pm}0.017$ & $0.521{\pm}0.023$ & --- & --- & --- \\
    EEF        & $19.23{\pm}0.34$ & $0.815{\pm}0.006$ & $0.171{\pm}0.007$ & $22.64{\pm}0.26$ & $0.878{\pm}0.005$ & $0.107{\pm}0.005$ & $20.32{\pm}0.25$ & $0.858{\pm}0.005$ & $0.183{\pm}0.008$ \\
  Skeleton   & $19.58{\pm}0.34$ & $0.824{\pm}0.006$ & $0.162{\pm}0.007$ & $22.74{\pm}0.28$ & $0.878{\pm}0.005$ & $0.106{\pm}0.004$ & $21.02{\pm}0.38$ & $\mathbf{0.866{\pm}0.005}$ & $0.169{\pm}0.010$ \\
    Ours       & $\mathbf{22.90{\pm}0.52}$ & $\mathbf{0.842{\pm}0.007}$ & $\mathbf{0.123{\pm}0.007}$ & $\mathbf{23.74{\pm}0.24}$ & $\mathbf{0.887{\pm}0.004}$ & $\mathbf{0.095{\pm}0.004}$ & $\mathbf{22.79{\pm}0.35}$ &
   $0.864{\pm}0.008$ & $\mathbf{0.148{\pm}0.010}$ \\
  \bottomrule
  \end{tabular}%
  }
  \end{table}
For completeness, in Table~\ref{tab:recon-all} we present the quantitative results across all reconstruction experiments, along with the standard error for each metric. Our method consistently outperforms all the baselines, and within the standard error of conditioning on the skeleton for the real-world data we captured using an unseen custom gripper on the robot.

\section{Additional training details}
In DROID, to ensure that our validation and testing do not include any scenes from the training data, we exclude all scenes from the labs CLVR and RAD from the training set and reserve them for our test set. 
We use approximately 1,000 DROID demonstrations with the two external cameras and process each demonstration with both the segmentation-based and rendering-based pipelines. We ensure that we keep the failure cases from DROID to ensure that the model can generate counterfactuals accurately. 
We also include 4,000 examples from Robocasa across all tasks that do not involve robot navigation, and generate failure cases to incorporate in the training data. Since the base video model requires a prompt as a condition, we include only a high-level prompt to ensure the model relies solely on the visual conditioning signal for video generation.

\section{VLM Evaluation Protocol}

In the policy evaluation, we used an off-the-shelf VLM to assess the generated videos, fully automating the process. Our evaluation protocol is as follows:  
For each generated rollout video, we query Gemini~3.1~Pro Preview
(\texttt{gemini-3.1-pro-preview}, temperature $0$, sampled at 15  fps so all $81$ frames of the  clip are seen) with a task-agnostic system instruction that calls out three world-model failure modes, the evaluator must penalize: ghost/weak contact (end-effector near but not touching), post-disengagement
coasting (object continues toward the goal after the end-effector retracts), and
frame-jump glitches (goal state appears after a teleport). The model returns a
structured JSON object with four scored fields: \texttt{robot\_caused\_outcome}
(boolean \texttt{caused}, categorical \texttt{mechanism} $\in$ \{
\texttt{robot\_pushed}, \texttt{robot\_grasped}, \texttt{teleported},
\texttt{vanished}, \texttt{autonomous}, \texttt{passthrough}, \texttt{no\_attempt},
\texttt{other}\}), \texttt{task\_success} (boolean plus confidence),
\texttt{end-effector\_contact} (boolean plus target object), and \texttt{physics\_realism}
(1--5 score with an \texttt{issues} list). Field order is enforced via the response
schema so that the causal mechanism is committed \emph{before} the binary success
flag, anchoring the verdict on process rather than the final frame. Task-specific
success criteria (e.g.\ ``the toaster door must be flush, latched, and closed by a
push from below'') are passed in the per-rollout user prompt rather than encoded
in the system instruction.

To select one rollout per scene from $N_s$ candidates, we sort by the lexicographic
key $\kappa = (\mathbf{1}[\text{success}], \text{realism}, \text{confidence},
\mathbf{1}[\text{contact}], -r)$ in descending order: prefer success-flagged
rollouts; among them, prefer higher physics realism; then higher confidence; then
visible contact; tiebreak deterministically by lower rollout index $r$. The pick is
$\arg\max_r \kappa(v_{s,r})$, and the planner's per-task success rate is the
fraction of scenes for which the picked rollout is GT-correct
($\texttt{max\_reward}=1.0$). To plot success as a function of~$N$, we exhaustively
enumerate every size-$N$ subset of the scene's rollouts, take Gemini's top-ranked
rollout within each subset, and average the resulting hit rate across the
$\binom{N_s}{N}$ subsets and across scenes. The random baseline reduces to this
expression at $N{=}1$, where the planner has no choice and merely returns the
per-scene mean GT rate.

\subsection{Gemini system prompt}
\label{sec:gemini-prompt}

The system instruction below is sent verbatim with every video query (model: \texttt{gemini-3.1-pro-preview}, temperature $0$, sampled at 15 fps). 
Per-rollout user prompts add only the task description and the task-specific success criterion; all evaluation policy lives in this system instruction.

\begin{lstlisting}[style=prompt]
You are a strict evaluator of robot-manipulation videos.
The video was produced by a generative video model that is meant to
*simulate* a robot policy executing a task. The video is often
visually flawed and may HALLUCINATE the desired outcome -- e.g. the
target object teleports into its goal pose, vanishes and reappears in
place, or moves on its own without the robot causing it. Your job is
to judge whether the *robot* actually caused the outcome via
plausible physical contact, not whether the final frame happens to
look correct.

CRITICAL ANTI-PATTERNS -- these are common world-model artifacts that
look superficially like success but must NOT be credited as
robot_pushed / robot_grasped:

  (a) Ghost / weak contact. The gripper must be in visible mechanical
      coupling with the object on the frames the object moves: the
      gripper geometry should be touching/compressing against the
      object surface, and the object's motion should be co-located in
      time with that contact. If the gripper is merely *near* the
      object, in front of it, or has only momentary glancing contact
      and the object still moves, treat the motion as autonomous or
      passthrough -- NOT as robot_pushed. "Implied" or "magic" contact
      does not count.

  (b) Post-disengagement coasting / "momentum". The simulated
      environment has NO momentum or follow-through: an object does
      not continue moving toward its goal after the gripper retracts
      or stops. If the gripper disengages (retracts, stops, or moves
      away) and the object continues to move closed/into-goal on its
      own afterwards, the post-disengagement frames are autonomous,
      regardless of what came before. Only credit the extent of
      progress that occurred WHILE the gripper was in active contact
      AND moving the object. If the goal state is only reached after
      the gripper has disengaged, robot_caused_outcome.caused is
      FALSE and task_success.success is FALSE.

  (c) Glitchy reach-around or frame jumps. If the goal state appears
      in a frame following a visual discontinuity, scene reset, or
      sudden teleport -- even if the robot's pose looks plausible
      around it -- treat as teleported.

Score these fields:

  1. robot_caused_outcome
       caused      True only if the robot, via its gripper, is the
                   visible physical cause of the goal-relevant motion,
                   AND the goal state is reached DURING the contact
                   window (not after it).
       mechanism   one of:
                     "robot_pushed"      gripper pushed/pulled the object
                                         in sustained contact
                     "robot_grasped"     gripper grasped and moved it
                     "teleported"        object snapped to goal state
                     "vanished"          object disappeared from scene
                     "autonomous"        object moved on its own
                                         (incl. post-disengagement
                                         coasting, ghost contact)
                     "passthrough"       gripper went through the object
                     "no_attempt"        robot never engaged the object
                     "other"             describe in reasoning
  2. task_success      success=TRUE requires ALL THREE of: (a) the
                       end-state matches the success criterion, (b)
                       robot_caused_outcome.caused is true, and (c)
                       the goal state was reached during a frame
                       range where the gripper was actively in
                       mechanical contact with the object. Outcomes
                       reached after disengagement, via ghost contact,
                       teleport, or autonomous motion are FALSE.
  3. gripper_contact   Did the gripper visibly make physical contact
                       with the target object? "Hovering near",
                       "passing through", or single-frame glancing
                       contact does not count.
  4. physics_realism   1-5 plausibility. Penalize teleporting,
                       morphing, gripper passing through solids,
                       vanishing or duplicated objects, frame-to-frame
                       jumps, ghost contact, post-disengagement
                       coasting.

Be terse but specific. Reference what you see in the video (frame
ranges, objects, motions, whether the gripper was in contact when
the goal state was reached). Do not speculate beyond what is
visible.
\end{lstlisting}

\paragraph{Per-rollout user message.} On top of the system prompt, the per-video user message provides only the natural-language task and the success criterion:

\begin{lstlisting}[style=prompt]
Intended task: {task_prompt}
Success criterion: {expected_outcome}
Score physics_realism on a 1-5 integer scale: 1 = nonsensical,
3 = obvious artifacts but recognizable, 5 = indistinguishable from
real video. Return ONLY the JSON object matching the response schema.
\end{lstlisting}

\paragraph{Structured response schema.} The model is constrained to
return JSON with five top-level fields, in this order (enforced via
\texttt{propertyOrdering}): \texttt{robot\_caused\_outcome} (booleans
\texttt{caused}, categorical \texttt{mechanism}, free-form
\texttt{reasoning}), \texttt{task\_success} (\texttt{success},
\texttt{confidence}, \texttt{reasoning}), \texttt{gripper\_contact}
(\texttt{contact}, \texttt{target\_object}, \texttt{reasoning}),
\texttt{physics\_realism} (integer \texttt{score} $\in [1,5]$,
\texttt{issues} list, \texttt{reasoning}), and a free-form
\texttt{summary}. Ordering causal mechanism \emph{before} the binary
success flag forces the model to commit to a process explanation
rather than rationalize from the final frame.

\section{Policy and Inverse-Dynamics Training Details}

Here, we discuss the training details for policy learning and the inverse-dynamics model.
We distinguish between three settings:
\emph{(i)} simulation policies used for planning and policy evaluation in RoboCasa,
\emph{(ii)} the learned inverse-dynamics model used for action extraction, and
\emph{(iii)} policies trained for real-world rollout studies.

\subsection{Simulation Policy Training for Planning and Policy Evaluation}

Unless otherwise stated, the policy used in our planning and policy-evaluation experiments is a Diffusion Policy~\citep{chi2023diffusionpolicy, chi2024diffusionpolicy} trained on task-specific mixtures of human demonstrations and MimicGen~\cite{mandlekar2023mimicgen} trajectories provided by the official RoboCasa~\citep{robocasa365} sources.
Each policy conditions on RGB observations from two fixed cameras together with the robot state, and predicts the full $12$-D action sequence for the episode horizon. Concretely, these $12$ dimensions are $3$ end-effector position commands, $3$ end-effector rotation commands, $1$ end-effector-close command, $3$ mobile-base commands, $1$ torso command, and $1$ control-mode command.
Across tasks, the human teleoperation data contain approximately $100$ demonstrations, while the MimicGen augmentation expands the training set to roughly thousands of trajectories, depending on the task and horizon.
For example, in the MG+Human setting used for our atomic-task studies, the merged training sets are typically on the order of $9\times 10^3$ trajectories.

For the simulation policies, we evaluate them against the success conditions provided by the benchmark.
In planning, these policies serve as proposal generators: we sample action candidates from the policy, simulate each candidate with the video model, and then rank the predicted futures with the VLM-based evaluator.
In policy evaluation, the same class of policies provides the action trajectories whose success rate is compared between the simulator and the video model.

\subsection{Learned Inverse Dynamics for Action Extraction}

For action extraction, the video model first synthesizes a robot interaction conditioned on desired object motion, and a learned inverse-dynamics model then converts the generated video into an executable low-level action sequence.
This inverse-dynamics model is trained on \textsc{CoffeeServeMug} using the same native $12$-D RoboCasa action space.
The model predicts the full action sequence with a transformer-based architecture and is optimized with AdamW~\citep{loshchilov2017fixing} for $300$ epochs.

We also include Diffusion Policy~\citep{chi2023diffusionpolicy, chi2024diffusionpolicy}, ACT~\citep{zhao2023learning}, and SmolVLA~\citep{shukor2025smolvla} baselines in this experiment, but only as narrow controls for the action-extraction setting.
Those baselines operate on the first frame of the generated video rather than the full synthesized interaction, and therefore do not require a separate dedicated discussion outside this subsection.

\subsection{Real-World Policy Training}

For the real-world rollout studies, we additionally train image-conditioned policies directly on two captured manipulation tasks: \emph{mug on rack} and \emph{close cabinet}.
These datasets are sampled at $10$ Hz and contain $40$ trajectories for each task.
Each example provides four synchronized RGB views together with an $8$-D joint-and-end-effector action sequence.
For both tasks, we train Diffusion Policy~\citep{chi2023diffusionpolicy, chi2024diffusionpolicy} and ACT~\citep{zhao2023learning} policies under the same observation-action contract.

\section{Robot Data Collection}
\label{app:robot_data_collection}

We collect real-robot data using a Franka Panda manipulator equipped with a custom 3D printed compliant end-effector. Demonstrations are collected through teleoperation with a custom GELLO interface. During each rollout, the operator controls the robot from an initial scene state $s_0$, while we record synchronized robot states and RGB observations from both wrist-mounted and external cameras.

\subsection{Hardware Setup}

The sensing setup consists of two ZED Mini 2 cameras. 
One camera is mounted on the robot's wrist and provides an egocentric view near the end-effector. The second camera is mounted externally and provides a fixed third-person view of the workspace. Both cameras record RGB frames at $720$p resolution and $30$ Hz.

\subsection{Camera and Robot Frame Calibration}

We calibrate the camera and robot frames using a table-mounted AprilTag. The AprilTag remains fixed in the workspace throughout calibration. To estimate its pose in the robot base frame, we move the Franka through a set of diverse poses and capture images of the tag from the wrist camera. For each pose, Franka kinematics provide the end-effector frame ($T^B_E(q_t)$), and AprilTag detection provides the tag pose in the wrist-camera frame. These observations are used to estimate the wrist-camera hand-eye transform $T^E_{C^{\mathrm{wrist}}}$ and the AprilTag pose $T^B_A$ in the robot base frame.

The external camera is then calibrated by observing the same table-mounted AprilTag. Since $T^B_A$ is known from the wrist-camera calibration, the external-camera pose relative to the robot base can be computed from the detected tag pose:
\[
T^B_{C^{\mathrm{ext}}}
=
T^B_A
\left(T^{C^{\mathrm{ext}}}_{A}\right)^{-1}.
\]
Thus, the table-mounted AprilTag provides a shared reference frame between the robot, wrist camera, and external camera. We verify calibration quality by projecting the robot model into the camera images and checking their alignment with the observed frames.

\subsection{Trajectory Representation}

Each rollout is stored as a trajectory
\[
\tau =
\{(I^{\mathrm{ext}}_t, I^{\mathrm{wrist}}_t, q_t)\}_{t=1}^{T},
\]
where $I^{\mathrm{ext}}_t$ and $I^{\mathrm{wrist}}_t$ are the external and wrist RGB observations, and $q_t$ is the measured robot joint state.

All data are stored in HDF5 format. Each file corresponds to one rollout and contains RGB frames, joint states, timestamps, and calibration metadata. The calibration metadata includes
\[
K_{\mathrm{ext}},
\quad
K_{\mathrm{wrist}},
\quad
T^B_A,
\quad
T^B_{C^{\mathrm{ext}}},
\quad
T^E_{C^{\mathrm{wrist}}}.
\]
For experiments involving visual robot masks, the stored joint states are combined with the Franka kinematic model, custom end-effector geometry, and calibrated camera parameters to render the robot geometry into the image plane.

\section{Real World Tasks Rubrics}
We score each demonstration by awarding one point per completed sub-stage, and report
\emph{task progress} as the fraction of points earned, so that scores are comparable
across tasks of differing length. The per-task rubrics are as follows.

\begin{itemize}
    \item \textbf{Bagging orange} (2 points): picking up the orange (1 point), and
          placing the orange inside the plastic bag (1 point).
    \item \textbf{Stacking blocks} (4 points): picking up the purple block (1 point),
          placing it on top of the orange block (1 point), picking up the black block
          (1 point), and placing it on top of the purple block (1 point).
    \item \textbf{Placing towel} (2 points): picking up the towel (1 point), and
          placing it inside the bowl (1 point).
    \item \textbf{Closing drawer} (1 point): fully closing the drawer.
\end{itemize}

\section{Adapting Baselines to Unseen Embodiments}
To evaluate generalization to unseen embodiments, we evaluate our model and baseline methods in BEHAVIOR-1K~\cite{li2024behavior1k}, where a bimanual humanoid robot is tasked to complete household mobile manipulation tasks.
Since our video model simply takes a masked visual action as input, adapting our model to unseen embodiments that are significantly different from what it has observed is straightforward, as we pass in the rendering of the new robot's URDF.
In Ctrl-World, since it is trained to predict multi-view video outputs, we adapt the camera view settings in BEHAVIOR to match DROID's setup.
For action conditioning, since Ctrl-World is trained on single-arm robots from DROID, while BEHAVIOR data uses a bimanual robot with a different action space, we first use a heuristic to identify the active arm based on the magnitude of the relative end-effector transformation. 
Then, we use the identified arm as the action conditioning for Ctrl-World. 
We find this to be an effective heuristic, as a majority of robot behaviors in the dataset consist of independent arm motions.
For the model conditioned on the visualization of the end effector pose, we include both end effectors in the conditioning video. Similarly, for the model conditioned on the robotic arm skeleton, we visualize the skeletons of each robot arm from the bimanual robot.


\end{document}